\documentclass[letterpaper, 10 pt, conference]{ieeeconf}  

\IEEEoverridecommandlockouts                              

\usepackage{graphicx} 
\usepackage{amsmath} 
\usepackage{amssymb}  
\usepackage{bm}
\usepackage[ruled]{algorithm2e}
\usepackage{algpseudocode}
\usepackage{textcomp}
\usepackage{lipsum}
\usepackage[table]{xcolor}
\usepackage{array,colortbl}
\usepackage[noadjust]{cite}
\usepackage{hyperref}
\hypersetup{
    colorlinks=true,
    linkcolor=blue,
    filecolor=magenta,      
    urlcolor=cyan,
    }

\usepackage{xcolor}

\usepackage{array}
\newcolumntype{P}[1]{>{\centering\arraybackslash}p{#1}}

\title{\LARGE \bf
GelSight Svelte: A Human Finger-shaped Single-camera Tactile Robot Finger with Large Sensing Coverage and Proprioceptive Sensing
}

\author{Jialiang Zhao$^{1}$ and Edward H. Adelson$^{1}$
\thanks{$^{1}$Computer Science and Artificial Intelligence Lab, Massachusetts Institute of Technology
    {\tt\small \{alanzhao, adelson\}@csail.mit.edu}}
}

\begin{document}

\maketitle
\thispagestyle{empty}
\pagestyle{empty}

\begin{abstract}

Camera-based tactile sensing is a low-cost, popular approach to obtain highly detailed contact geometry information. 
However, most existing camera-based tactile sensors are fingertip sensors, and longer fingers often require extraneous elements to obtain an extended sensing area similar to the full length of a human finger.
Moreover, existing methods to estimate proprioceptive information such as total forces and torques applied on the finger from camera-based tactile sensors are not effective when the contact geometry is complex.
We introduce GelSight Svelte, a curved, human finger-sized, single-camera tactile sensor that is capable of both tactile and proprioceptive sensing over a large area.
GelSight Svelte uses curved mirrors to achieve the desired shape and sensing coverage.
Proprioceptive information, such as the total bending and twisting torques applied on the finger, is reflected as deformations on the flexible backbone of GelSight Svelte, which are also captured by the camera.
We train a convolutional neural network to estimate the bending and twisting torques from the captured images.
We conduct gel deformation experiments at various locations of the finger to evaluate the tactile sensing capability and proprioceptive sensing accuracy.
To demonstrate the capability and potential uses of GelSight Svelte, we conduct an object holding task with three different grasping modes that utilize different areas of the finger.

More information is available on our website: \url{https://gelsight-svelte.alanz.info}.
\end{abstract}

\section{Introduction}

The sense of touch has been shown to be vital for humans to infer shapes \cite{klatzky1987there}, identify objects \cite{klatzky1992stages}, and manipulate objects \cite{westling1984factors}.
Researchers from the robotics community have also trained robots to perform tasks with tactile manipulators, such as localization \cite{bauza2022tac2pose, zhao2023fingerslam}, shape inference \cite{Suresh21tactile}, cloth folding \cite{sunil2022visuotactile}, and shape insertion \cite{dong2021tactile}.

Over the years, researchers have developed tactile sensors that work with different sensing methods, such as resistance, capacitance, magnetic, barometric, and optical sensors.
We refer readers to \cite{kappassov2015tactile} for an in-depth review of different types of tactile sensors and their applications.
Compared to other sensing principles, camera-based tactile sensors have the advantage of providing high-resolution geometric information of the contact surface while keeping the cost low.

However, most existing camera-based tactile sensors have a flat sensing area located only at the tip of the gripper.
Human fingers are long and round, and the sense of touch on the entire finger is important for their dexterity. 
Building a camera-based tactile sensor that has a similar shape and sensing area to human fingers is challenging.
Optically, the camera's field of view needs to cover a large sensing area that spans the entire length of the finger.
Mechanically, the shape of the finger needs to be similar to a human finger while containing the entire imaging and illuminating mechanisms.

We introduce GelSight Svelte, a human finger-sized tactile sensor that has a large sensing coverage along the entire finger.
The main idea behind GelSight Svelte is to achieve the desired finger curvature and sensing coverage by re-distributing the light using curved mirrors. 
Furthermore, the deformation of GelSight Svelte's flexible backbone is tracked from the camera view, and it is then processed to calculate proprioceptive information during manipulation, such as bending and twisting torques.
The finger design and the fabrication process is detailed in Sec. \ref{sec:design}.
The geometrical tactile sensing capability is demonstrated with a screw pressing task in Sec. \ref{sec:geo_sensing}.
The proprioceptive sensing capability is discussed and evaluated in Sec. \ref{sec:pro_sensing}.
A sample object holding task conducted with three GelSight Svelte fingers in different grasping modes is demonstrated in Sec. \ref{sec:obj_grasping}.

\begin{figure}[ht]
\centering
\includegraphics[width=\linewidth]{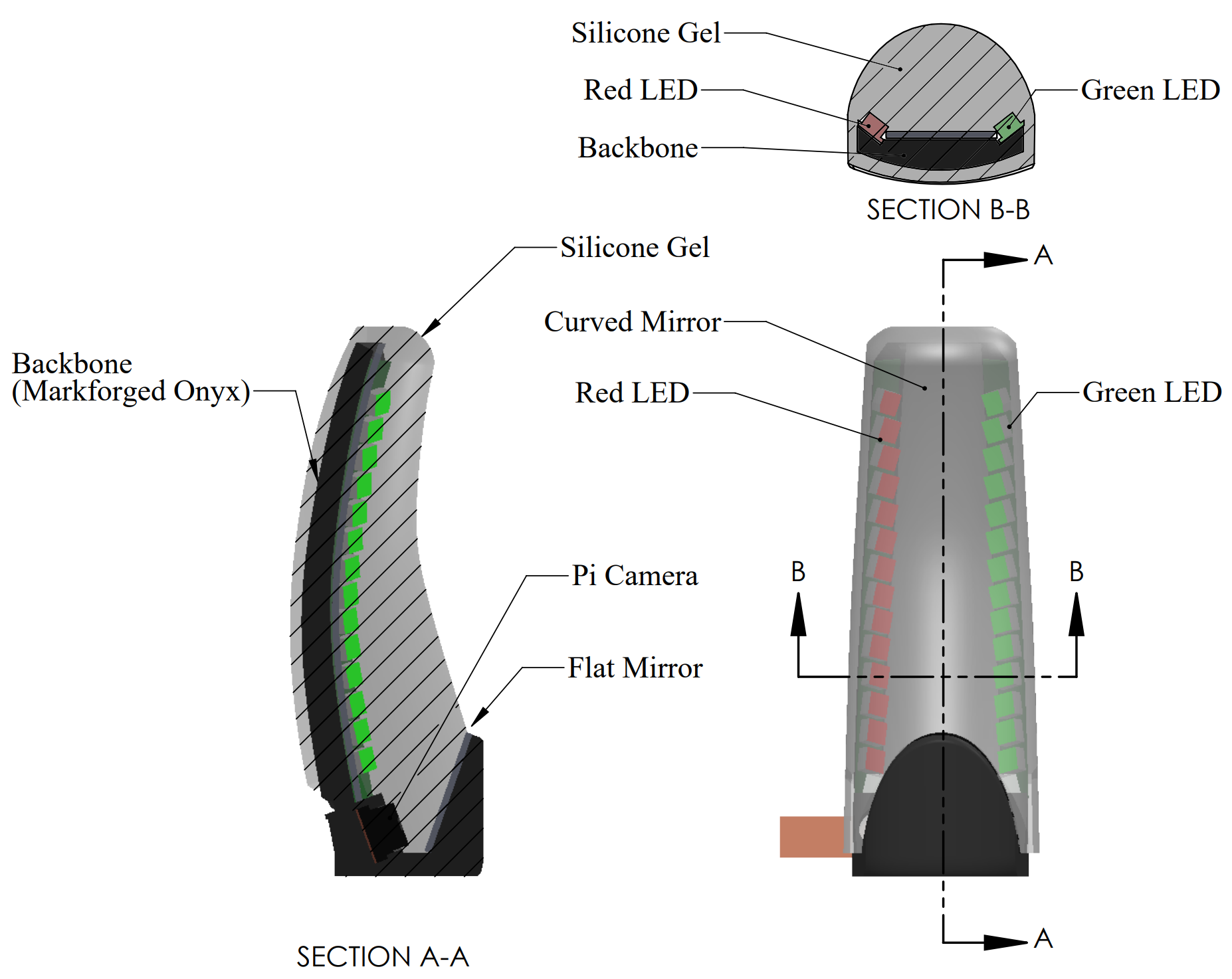}
\caption{Construction of a GelSight Svelte finger and section views. GelSight Svelte measures 83.5mm long, 22.7mm wide, and on average 18.8mm thick.}
\label{construction}
\end{figure}

\begin{figure*}[ht]
\centering
\includegraphics[width=\linewidth]{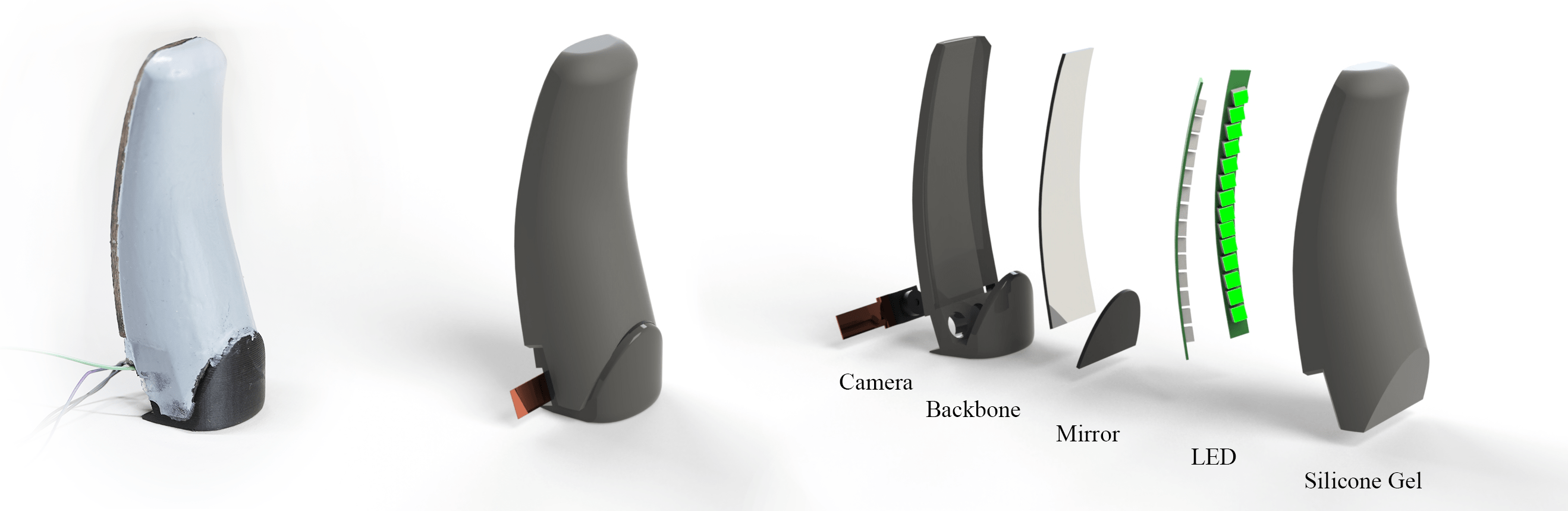}
\caption{\textit{Left:} a fully assembled GelSight Svelte finger. \textit{Middle: } a rendered assembled finger. \textit{Right:} a rendered exploded view. A GelSight Svelte finger is composed of 5 components: a camera, a backbone, two pieces of mirror, two LED strips, and a silicone gel.}
\label{exploded}
\end{figure*}




\section{Related Works}

Camera-based tactile sensors have the advantage of producing the geometry of the contact surface with high spatial resolution while keeping the form factor small and the cost low. 
One popular example of camera-based tactile sensors is the GelSight sensor, which relies on carefully designed LED illumination and the photometric stereo algorithm to reconstruct the 3D shape of the contact surface \cite{yuan2017gelsight}.
However, due to physical constraints such as size of the camera, focal distance of the lens, and viewing angles, most camera-based tactile sensors have the limitation of being flat, small and they lack a wide-range proprioceptive sensing ability.

\subsection{Configurations of camera-based tactile sensors}

One of the main limitations of camera-based tactile sensors is their shape. 
The geometrical requirements of illumination and the imaging path have typically led to sensors that are compact and/or flat. 
While these sensors can serve well as fingertips, they do not provide an extended sensing area similar to the full length of a human finger. 
Examples of flat or nearly flat sensors are given in \cite{yuan2017gelsight, wang2021gelsight, taylor2022gelslim, romero2020soft}. 
Rounded sensors are shown by \cite{tippur2022design, sun2022soft, padmanabha2020omnitact}. 
She et al. showed an elongated finger using an exoskeleton and a pair of cameras \cite{she2020exoskeleton}.
The exoskeleton and protruding cameras of this design limit its ability to be incorporated into compact grippers. 
Liu and Adelson showed a Fin Ray structure finger with camera-based sensing \cite{liu2022gelsight}. 
This finger has an extended sensing surface, but it is flat rather than rounded like a human finger. 
Wilson et al. designed a two-finger claw with an extended sensing area \cite{wilson2020design}, but the multiple sensing surfaces are all flat.

Our goal is to construct a finger that is similar in size to a human finger, with an extended sensing surface that is rounded, similar to the front surface of a human finger. 
We want to avoid extraneous elements such as exoskeletons and protruding cameras. 
To minimize excess bulk, our design uses a single camera that is completely housed within the finger. 
In order to satisfy our multiple constraints, we employ a folded optical path that uses two mirrors, where one of the mirrors is curved.

\subsection{Capturing proprioceptive information}

In addition to the local information provided by tactile sensing, it is often useful to have information about the overall forces, torques, and deformations associated with a finger. 
While it is often possible to infer such measurements from local tactile measurements, it may require significant computation. 
Moreover, the viscoelastic behavior of soft elastomers puts limits on the accuracy and repeatability of these estimates. 
In the Fin Ray tactile sensor introduced in \cite{liu2022gelsight}, Liu and Adelson printed dots on the Fin Ray structure, which is then tracked from the camera view to estimate the finger's proprioceptive state.
Building upon a similar concept, we construct our finger with a flexible backbone, and by measuring the flexing of this backbone we can estimate the bending and twisting torque being experienced by the finger.

\section{Design and Fabrication}
\label{sec:design}

GelSight Svelte utilizes two mirrors to increase sensing coverage and a flexible backbone to obtain proprioceptive information.
High-resolution tactile signals and proprioceptive information can be obtained from multiple surfaces on the sensor.
GelSight Svelte measures 83.5mm long, 22.7mm wide, and on average 18.8mm thick.

\subsection{Mechanical Design}

The construction of GelSight Svelte is illustrated in Fig.\ref{construction}, and an exploded view rendering is shown in Fig.\ref{exploded}.
GelSight Svelte is constructed with 5 components: a camera, a flexible backbone, two pieces of mirrors, two LED strips, and a silicone gel.

\begin{itemize}
    \item \textbf{Camera.} Raspberry Pi 120 degree wide angle cameras are selected due to their small footprint, widely adjustable focus range, high resolution and ease of communication.
    This camera uses a 5MP 1/4 inch OV5647 CMOS sensor, and the maximum resolution for video streaming is 1080P.

    \item \textbf{Backbone.} The finger backbone is 3D printed from Markforged Onyx, a carbon fiber-filled nylon composite material. This material combines high tensile strength and medium elasticity.

    \item \textbf{Mirror.} The two pieces of mirrors are cut out from 1mm thick flexible acrylic plastic sheet mirrors with a laser cutter. The curved piece of mirror is then bent into the desired shape.

    \item \textbf{LED strips.} We choose red and green Chanzon 3528 lensless LEDs for illumination. Each LED strip consists of 18 LEDs and a flexible PCB.

    \item \textbf{Silicone Gel.} We choose Silicone Inc. XP-565 silicone for its optical transparency and softness.
\end{itemize}
 
Note that the silicone gel encloses a large part of the entire assembly to increase the resistance to delamination.

\begin{figure}[ht]
\centering
\includegraphics[width=0.8\linewidth]{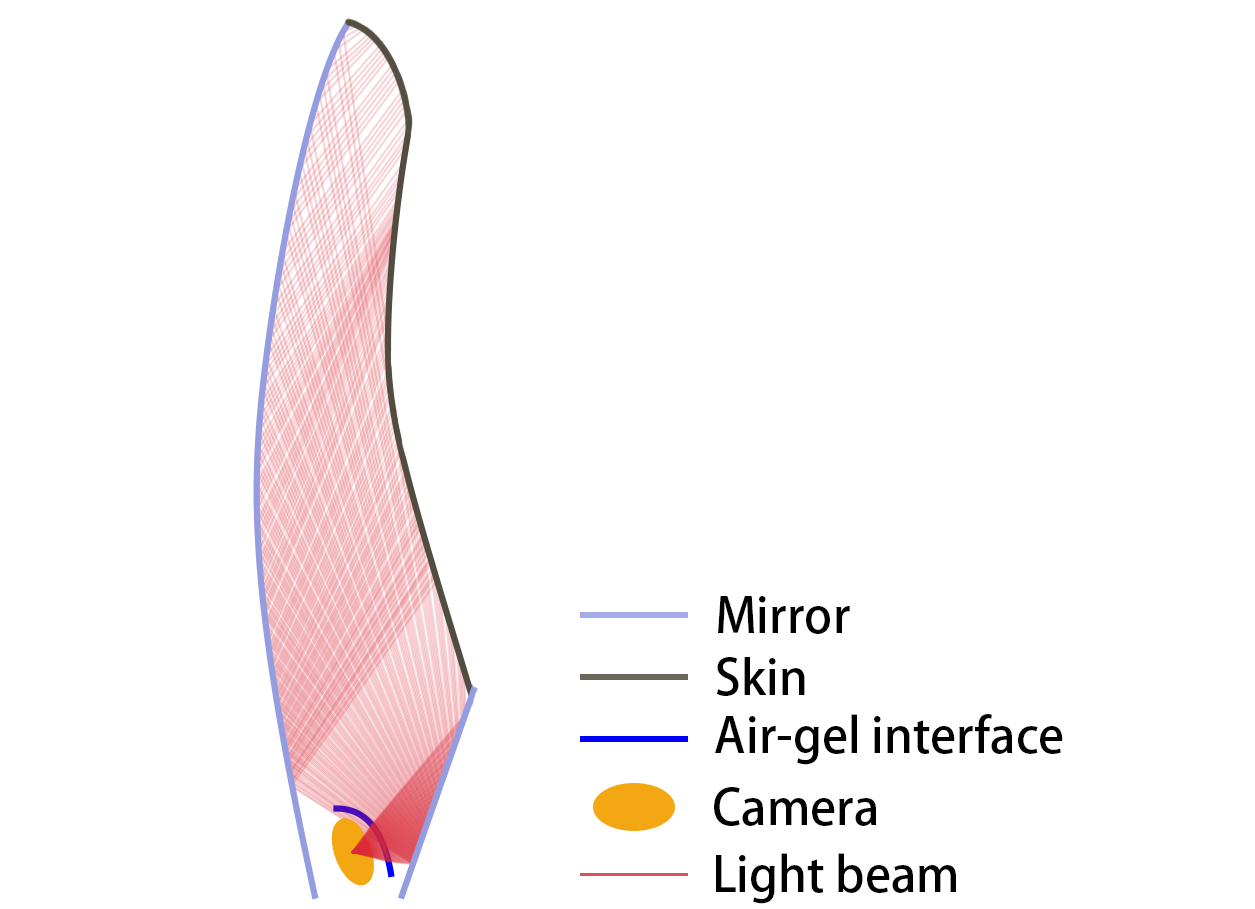}
\caption{Reflection / refraction simulation of GelSight Svelte. The curved mirror (light blue, left) and the sensing skin (gray) are modeled as B-splines. The air-gel interface (blue) is a hemisphere. The finger is designed so that the imaging angles (the angle between the red light beams and the gray sensing skin) are not too small and the entire sensing skin is covered by the camera's field of view (the red light beams distribute on the entire gray sensing skin).}
\label{ray_trace}
\end{figure}

\subsection{Optics and illumination design}

The curvatures of the mirror and the sensing skin are designed so that the camera's field of view should cover the entire sensing skin, and the overall dimensions should be close to human finger dimensions.
To achieve those goals, the sensing skin curvature and the mirror curvature are modeled as B-splines, and a reflection / refraction simulator was created to simulate the coverage and imaging angles.
Note that refraction happens at the air-gel interface near the camera.
The refraction index of the silicone gel we used is around $1.41$.
In our experiment, a flat air-gel interface leads to a very narrow field of view of the camera (around 90 degrees, versus 120 degrees when placed in the air).
To minimize this issue, a dome-shaped air-gel interface was designed.
A reflection / refraction simulation of the final curvature design is shown in Fig. \ref{ray_trace}.

\begin{figure}[ht]
\centering
\includegraphics[width=\linewidth]{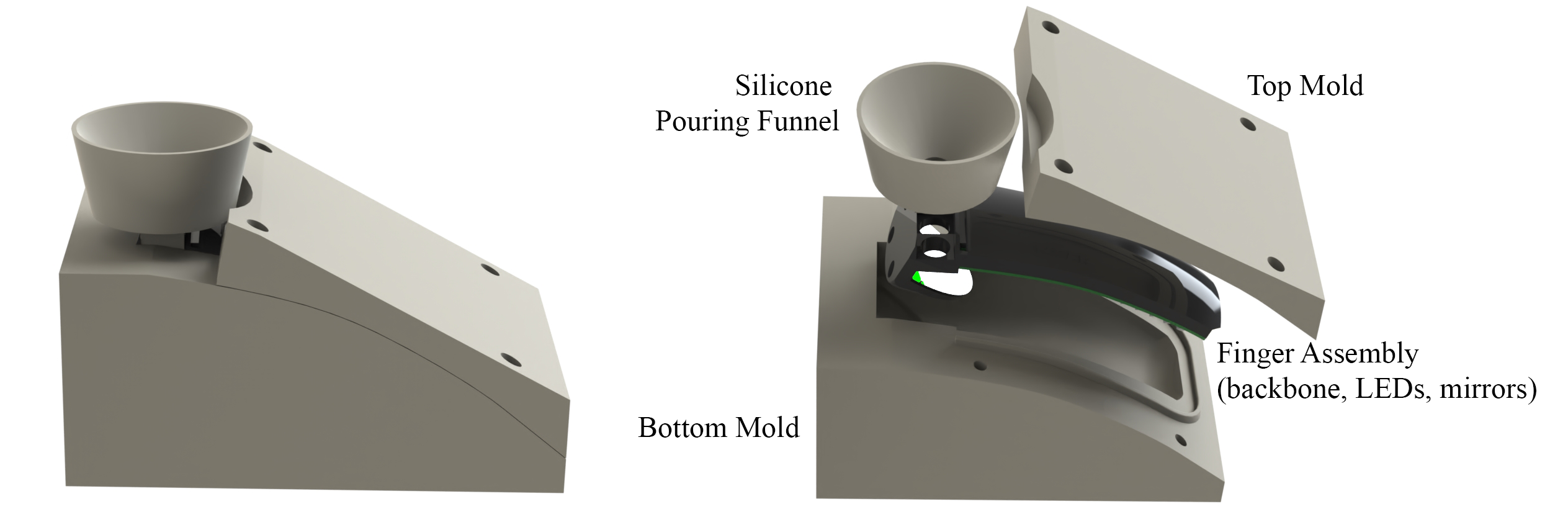}
\caption{The molding process. Two pieces of molds are used to cast silicone: a bottom mold and a top mold. A silicone pouring funnel is designed for easy pouring. }
\label{mold}
\end{figure}

\begin{figure}[ht]
\centering
\includegraphics[width=\linewidth]{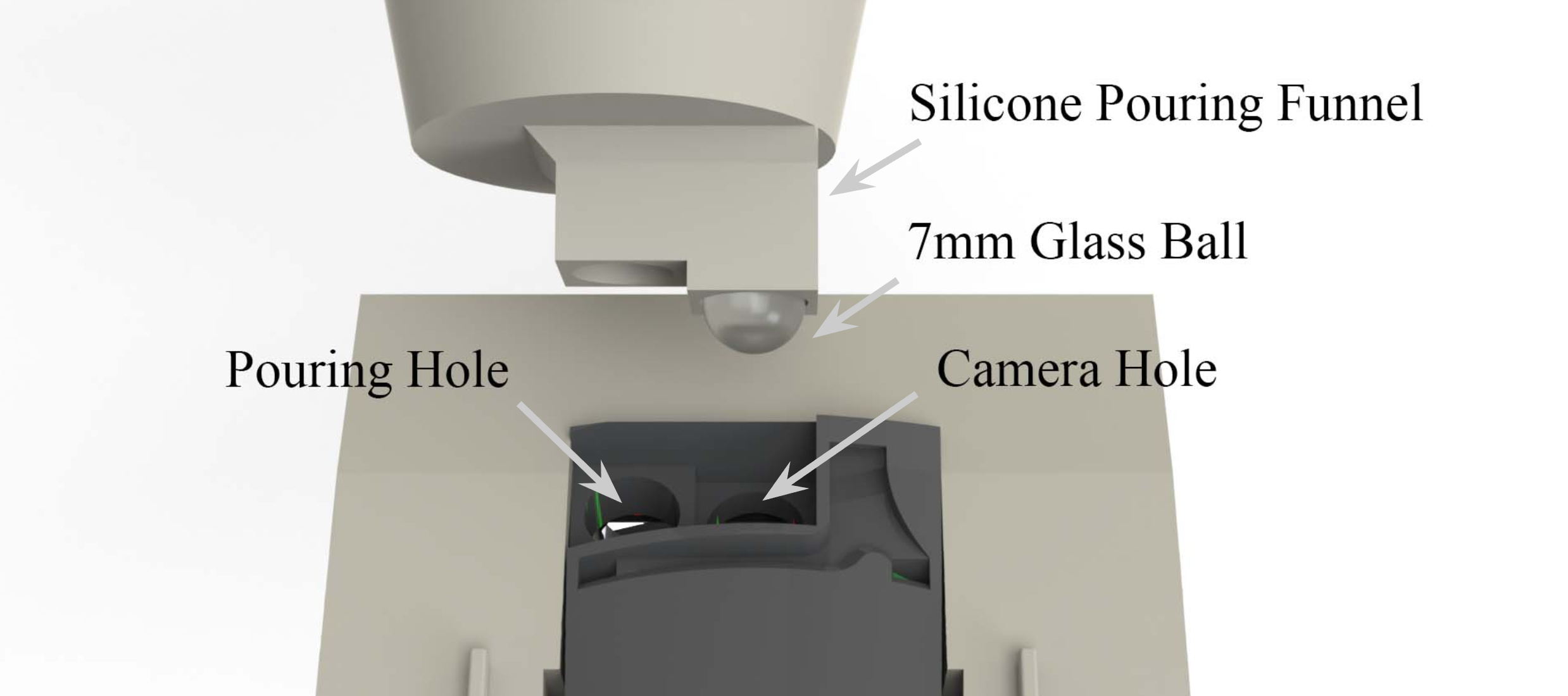}
\caption{A 7mm glass ball is glued to the bottom of the silicone pouring funnel to create the hemispherical air-gel interface. Liquid silicone flows from the funnel into the mold, which constitutes the body of the finger after curing.}
\label{pour}
\end{figure}

\subsection{Fabrication}

\subsubsection{Backbone and mold preparation}
The flexible backbone is 3D printed with Markforged Onyx, which is a carbon-fiber filled nylon composite filament.
Two pieces of molds are designed for the silicone casting process, as shown in Fig. \ref{mold}.
The mold pieces are 3D printed from ABS.
In order to make the cast silicone sensing surface smooth, the ABS mold pieces are first polished with sand paper, then further smoothed with pure acetone.

\subsubsection{LED strips and mirrors preparation}
One green LED strip and one red LED strip are used for illumination.
Each strip has 18 Chanzon 3528 LEDs on a 0.2mm thick flexible PCB.
The LED strips are then glued onto both side of the backbone.
The mirrors are laser cut from 1mm thick flexible acrylic mirror sheets, then glued onto the backbone.

\subsubsection{Plastic-silicone adhesion}
In order to avoid delamination between the plastic backbone and the silicone gel due to shear, we apply silicone adhesive to the finger assembly before casting silicone onto it.
We chose A-564 medical silicone adhesive, diluted with NOVOCS silicone solvent with a 1:1.8 ratio. 
After the LED strips and the mirrors are glued onto the backbone, we dip the assembly into the mixture and let the excess mixture fully drip off the finger.

\subsubsection{Reflective paint}
We use the following formula to make the gray reflective paint: 2g PRINT-ON gray silicone ink, 0.2g catalyst, 0.25g $6\mu m$ aluminum flake, and 6g NOVOCS silicone solvent.
After thoroughly mixing all the components, we spray the mixture onto the bottom mold with an airbrush.

\begin{figure*}[t!]
\centering
\includegraphics[width=\linewidth]{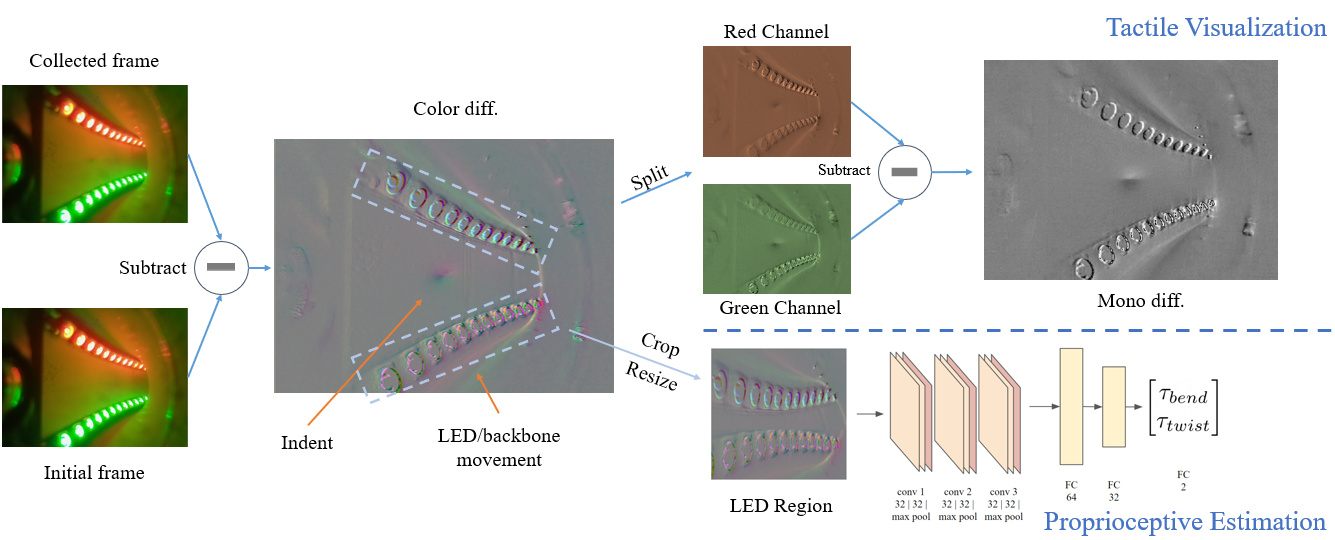}
\caption{Image preparation. First, a color difference image is created by subtracting an initial frame (captured when no object is pressing the finger) from each collected frame. (1) For tactile visualization, a monochrome difference image is created by subtracting the green channel from the red channel of the color difference image. (2) For proprioceptive estimation, we first crop and resize the color difference image so that only the two LED regions are kept, then we train a convolutional neural network to estimate the bending and twisting torques.}
\label{data_prep}
\end{figure*}

\subsubsection{Casting}
The silicone casting process involves two mold pieces, a silicone pouring funnel, and a finger assembly (assembled backbone, two mirror pieces, and two LED strips).
The casting assembly is illustrated in Fig. \ref{mold}.
In order to create the hemispherical air-gel interface, a 7mm glass ball is glued onto the bottom of the silicone pouring funnel, as shown in Fig. \ref{pour}.
We mixed Silicone Inc. XP-565 and catalyst 18:1 for the silicone material.
The mixture is then poured into the mold from the pouring funnel.

\section{Geometrical and Proprioceptive Sensing}

GelSight Svelte is able to measure the geometry of the contact surface, and provide proprioceptive information such as torques applied to the finger.
We conduct one experiment for each task, and we design a object holding task done with different grasping modes to demonstrate the potential uses for GelSight Svelte.
The image preparation workflow for all tasks are illustrated in Fig. \ref{data_prep}.
We create monochrome difference images to visually inspect the tactile images, and we train a neural network to predict the bending and twisting torques from the LED strips movement.

\subsection{Geometrical sensing}
\label{sec:geo_sensing}

\begin{figure}[ht]
\centering
\includegraphics[width=\linewidth]{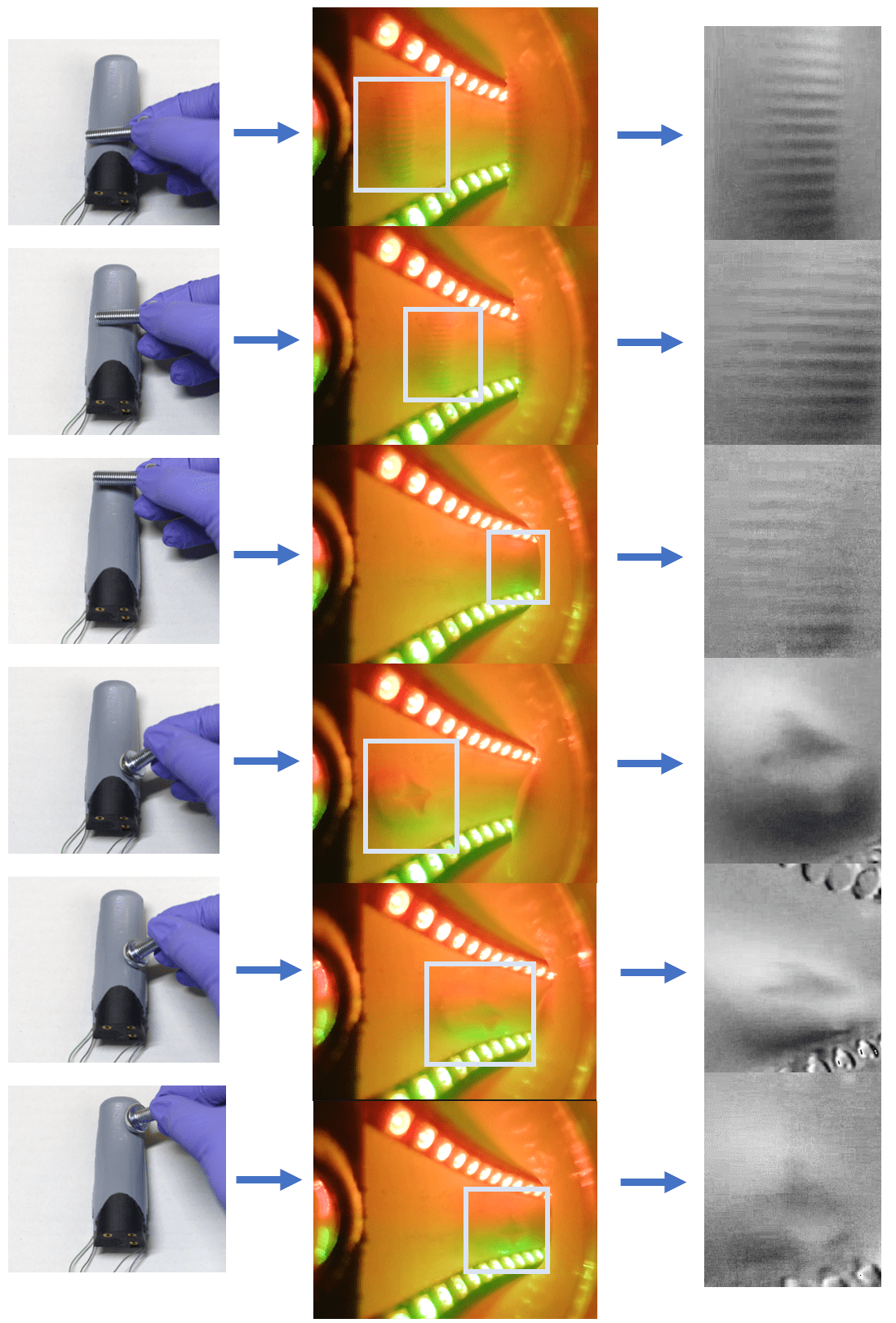}
\caption{Screw pressing task for geometrical sensing evaluation. A M8-1.25 screw is pressed at different locations of the finger. \textit{Left: } screw pressing at different locations. \textit{Middle: } raw tactile images. \textit{Right: } monochrome difference images zoomed at the contact locations.}
\label{pressing}
\end{figure}

We qualitatively examine the geometrical sensing capability of GelSight Svelte by pressing an object with known geometry at different locations of the finger.
A M8-1.25 screw (8mm diameter, 1.25mm thread pitch) is used for this task.
The experiment process and the collected tactile images are shown in Fig. \ref{pressing}.
The screw head as well as the threads are visible at all pressing locations, which means GelSight Svelte is able to provide reasonable resolution at all designed sensing areas. 
However, the quality differs at different locations, which is reflected as color saturation and contrast changes.
Higher saturation and contrast mean the image contains more information about the height gradients of the contact surface.
The contrast is the highest at the middle section of the finger, and the lowest at the tip of the finger.
This is due to the fact that the camera looks at the finger tip at a lower angle, i.e. less vertical to the sensing skin.
Although the light travelling distance at the fingertip is greater than the light travelling distance at the bottom of the sensing skin, the out-of-focus issue is insignificant.
The spatial resolution at maximum video resolution ($1920\times 1080\ px$) is around $30\ px/mm$ at the bottom of the sensor, and $19\ px/mm$ at the tip of the sensor.

\subsection{Proprioceptive sensing}
\label{sec:pro_sensing}
While it is possible to obtain proprioceptive information by printing and tracking markers on the finger skin, this method is usually only effective for local proprioceptive sensing, such as local normal force and local shear.
For global proprioceptive information such as the global force and torque applied on the finger, this method is less accurate due to the fact that the marker movement is highly related to the contact texture.
The sensible range of this method is also limited by the thickness and stiffness of the chosen silicone elastomer.

\begin{figure}[ht]
\centering
\includegraphics[width=\linewidth]{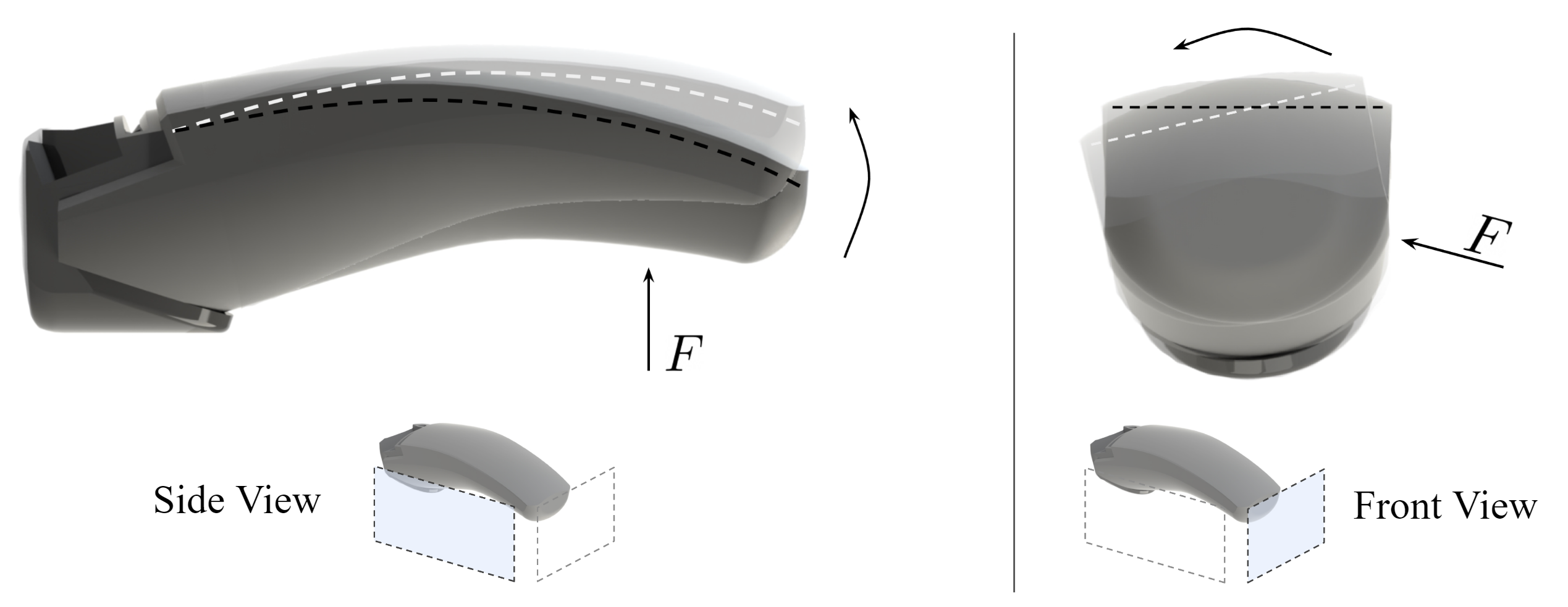}
\caption{Bending and twisting. \textit{Left: } (side view) when an upward vertical force is applied to the finger, the finger bends up. \textit{Right: } (front view) when a force is applied on the side of the finger, the finger twists.}
\label{bending}
\end{figure}

The backbone of GelSight Svelte is made from high tensile strength flexible materials.
When external pressure is applied, the finger bends or twists within a small range.
An illustration of the bending and twisting process is shown in Fig. \ref{bending}.
It is possible to estimate the amount and direction of the net external pressure applied onto the finger by tracking the movement of the finger backbone.
Because the LED strips of GelSight Svelte are fixed onto the backbone, we can track the backbone movement by tracking the LED movement from the camera view.

We collect tactile images when applying different amount of pressure at different locations of the finger.
A convolutional neural network is then trained to estimate the bending torque and the twisting torque from the LED movement.

\begin{figure}[ht]
\centering
\includegraphics[width=\linewidth]{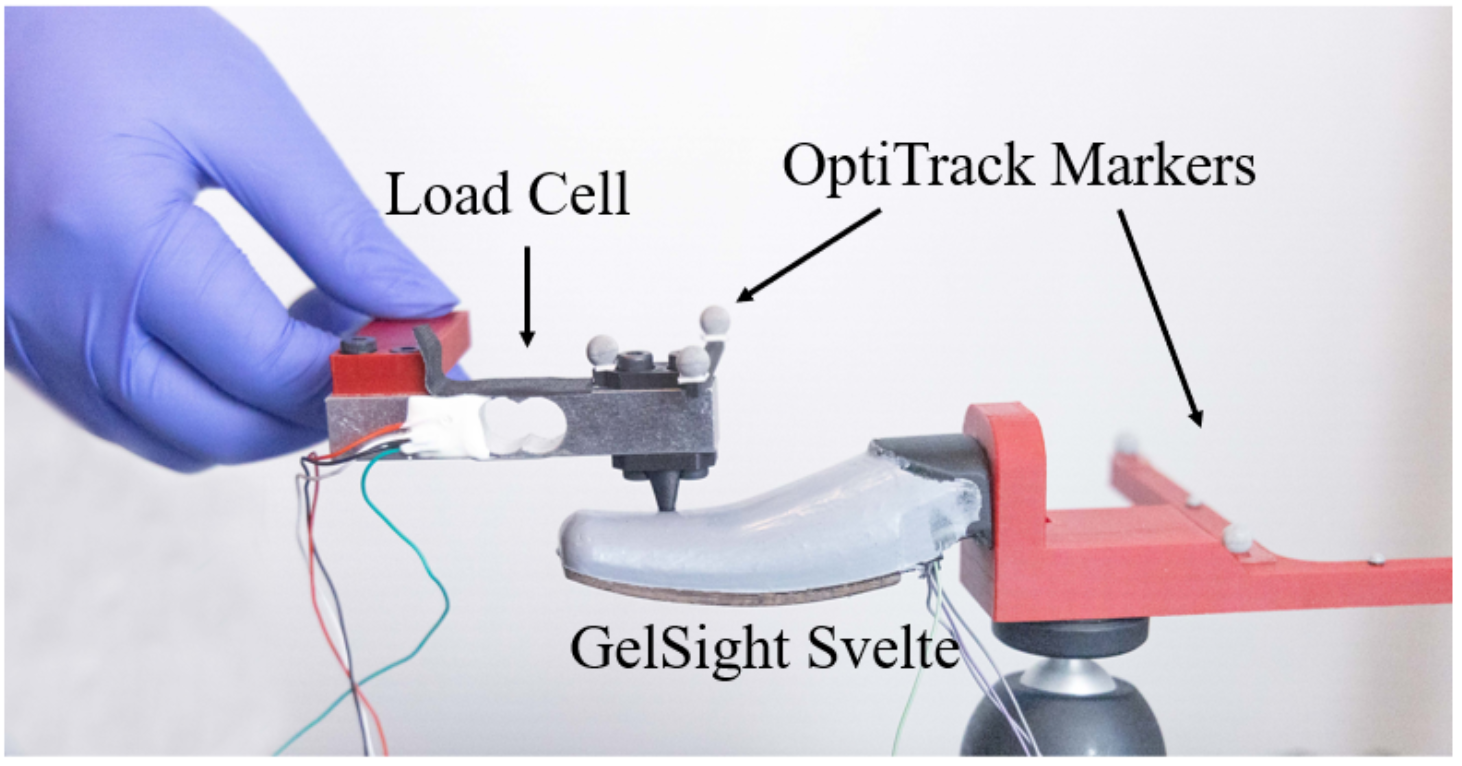}
\caption{Data collection for proprioceptive sensing. A human operator presses a load cell probe against a GelSight Svelte finger. The force is recorded from the load cell, and the relative transformation between the load cell and the finger is capture by an OptiTrack motion capture system. Bending torque and twisting torque are calculated from the force and the transformation.}
\label{data_collection}
\end{figure}

\begin{itemize}
    \item \textit{Data collection.} We mount a GelSight Svelte finger at a known location. We then probe the finger with a load cell, which accurately measures the applied force. The location of the finger and the location of the load cell are tracked in real time with an OptiTrack motion capture system. We then use the force measurement and the relative position between the finger and the load cell to calculate the bending torque and the twisting torque applied on the finger. Tactile images are collected. This data collection setup is shown in Fig. \ref{data_collection}.
    
    \item \textit{Image pre-processing and data augmentation.}
    We obtain the "difference image" by subtracting the initial frame (captured when no object is pressing the finger) from the collected frame.
    Because only the LED movements are relevant to proprioceptive sensing, we crop the images and only keep the two regions that contain the two LED strips. To achieve better generalization across hardware, we apply a randomly sampled scaling and shifting factor to each image. This process is shown in Fig. \ref{data_prep}.

    \item \textit{Neural network training.}
    A convolutional neural network is constructed and trained to estimate the bending and twisting torques from the processed images. The network architecture is shown in Fig. \ref{data_prep}.
\end{itemize}



\begin{figure}[ht]
\centering
\includegraphics[width=\linewidth]{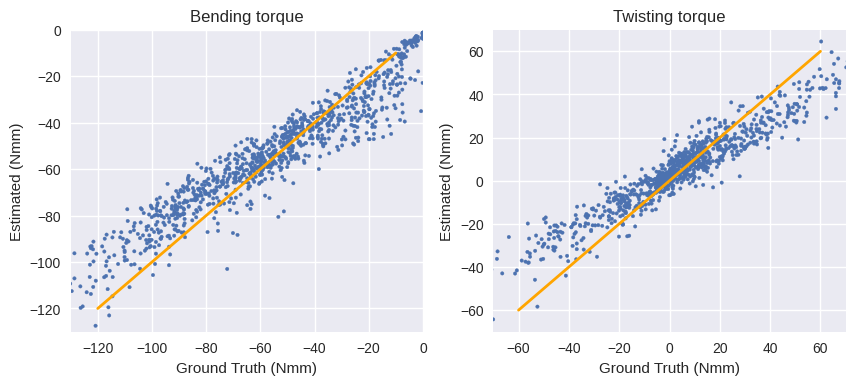}
\caption{Distributions of bending and twisting torques estimations.}
\label{scatter}
\end{figure}

We evaluate the torque estimation accuracy by performing the same probing task on a different finger that was not used in the data collection process.
The Root Mean Square Error (RMSE) for the estimated bending torque is 9.4 Nmm, where the mean and standard deviation across the training data are -62.4 Nmm and 28.3 Nmm, respectively.
RMSE for the estimated twisting torque is 7.6 Nmm, with a mean of 5.4 Nmm and standard deviation of 27.3 Nmm in the training data.
Distributions of the estimation are shown in Fig. \ref{scatter}.

\begin{figure*}[ht]
\centering
\includegraphics[width=0.95\linewidth]{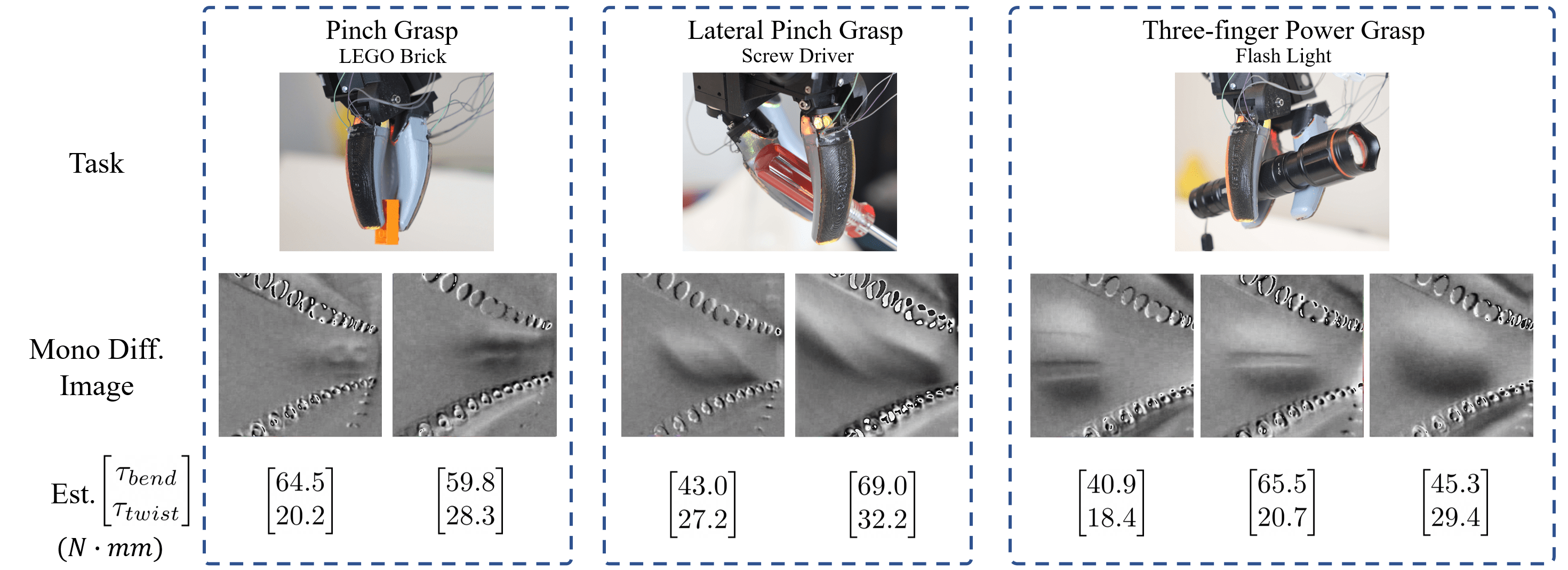}
\caption{The object holding experiment. \textit{Left: } holding a LEGO brick with a two-finger pinch grasp. \textit{Middle: } holding a screw driver with a two-finger lateral pinch grasp. \textit{Right: } holding a flash light with a three-finger power grasp.}
\label{picking}
\end{figure*}

\subsection{Object holding}
\label{sec:obj_grasping}
Having large contact area robotic manipulators enables easier and more stable holding of objects or tools.
We design a three-finger robot hand that's capable of pinch grasps, lateral pinch grasps, and three-finger power grasps using GelSight Svelte fingers.
We conduct three object picking tasks using each of the grasping mode in a open-loop manner, and we show the collected monochrome difference images and estimated torques for each task.
The three tasks are:
\begin{itemize}
    \item picking up a LEGO brick with antipodal pinch grasps. Two fingers were used.
    \item picking up a screw driver with lateral pinch grasps. Two fingers were used.
    \item picking up a flash light with three-finger power grasps. Three fingers were used.
\end{itemize}
We show task illustrations, collected tactile images, and predicted torques in  Fig. \ref{picking}.

\section{CONCLUSIONS}

In this paper we present GelSight Svelte, a curved, human finger-shaped tactile manipulator that is capable of both tactile and proprioceptive sensing over a large sensing area.
We introduce two novel approaches for camera-based tactile manipulator design: (1) using curved mirrors to achieve large sensing area while keeping the form factor small, and (2) obtaining proprioceptive information by tracking the deformation of a flexible backbone.
The design and fabrication process is introduced in detail, and the sensing capability is discussed.








\section*{ACKNOWLEDGMENT}

Toyota Research Institute provided funds to support this work.
The authors thank Sandra Q. Liu and Megha Tippur for providing insights and assistance in the design and fabrication processes.


\bibliographystyle{IEEEtran}
\bibliography{refs}

\begin{thebibliography}{10}
\providecommand{\url}[1]{#1}
\csname url@rmstyle\endcsname
\providecommand{\newblock}{\relax}
\providecommand{\bibinfo}[2]{#2}
\providecommand\BIBentrySTDinterwordspacing{\spaceskip=0pt\relax}
\providecommand\BIBentryALTinterwordstretchfactor{4}
\providecommand\BIBentryALTinterwordspacing{\spaceskip=\fontdimen2\font plus
\BIBentryALTinterwordstretchfactor\fontdimen3\font minus
  \fontdimen4\font\relax}
\providecommand\BIBforeignlanguage[2]{{%
\expandafter\ifx\csname l@#1\endcsname\relax
\typeout{** WARNING: IEEEtran.bst: No hyphenation pattern has been}%
\typeout{** loaded for the language `#1'. Using the pattern for}%
\typeout{** the default language instead.}%
\else
\language=\csname l@#1\endcsname
\fi
#2}}

\bibitem{klatzky1987there}
R.~L. Klatzky, S.~J. Lederman, and C.~Reed, ``There's more to touch than meets
  the eye: The salience of object attributes for haptics with and without
  vision.'' \emph{Journal of experimental psychology: general}, vol. 116,
  no.~4, p. 356, 1987.

\bibitem{klatzky1992stages}
R.~L. Klatzky and S.~J. Lederman, ``Stages of manual exploration in haptic
  object identification,'' \emph{Perception \& psychophysics}, vol.~52, no.~6,
  pp. 661--670, 1992.

\bibitem{westling1984factors}
G.~Westling and R.~S. Johansson, ``Factors influencing the force control during
  precision grip,'' \emph{Experimental brain research}, vol.~53, pp. 277--284,
  1984.

\bibitem{bauza2022tac2pose}
M.~Bauza, A.~Bronars, and A.~Rodriguez, ``Tac2pose: Tactile object pose
  estimation from the first touch,'' \emph{arXiv preprint arXiv:2204.11701},
  2022.

\bibitem{zhao2023fingerslam}
J.~Zhao, M.~Bauza, and E.~H. Adelson, ``Fingerslam: Closed-loop unknown object
  localization and reconstruction from visuo-tactile feedback,'' \emph{arXiv
  preprint arXiv:2303.07997}, 2023.

\bibitem{Suresh21tactile}
S.~Suresh, M.~Bauza, K.-T. Yu, J.~G. Mangelson, A.~Rodriguez, and M.~Kaess,
  ``Tactile slam: Real-time inference of shape and pose from planar pushing,''
  in \emph{2021 IEEE International Conference on Robotics and Automation
  (ICRA)}.\hskip 1em plus 0.5em minus 0.4em\relax IEEE, 2021, pp.
  11\,322--11\,328.

\bibitem{sunil2022visuotactile}
N.~Sunil, S.~Wang, Y.~She, E.~Adelson, and A.~Rodriguez, ``Visuotactile
  affordances for cloth manipulation with local control,'' \emph{arXiv preprint
  arXiv:2212.05108}, 2022.

\bibitem{dong2021tactile}
S.~Dong, D.~K. Jha, D.~Romeres, S.~Kim, D.~Nikovski, and A.~Rodriguez,
  ``Tactile-rl for insertion: Generalization to objects of unknown geometry,''
  in \emph{2021 IEEE International Conference on Robotics and Automation
  (ICRA)}.\hskip 1em plus 0.5em minus 0.4em\relax IEEE, 2021, pp. 6437--6443.

\bibitem{kappassov2015tactile}
Z.~Kappassov, J.-A. Corrales, and V.~Perdereau, ``Tactile sensing in dexterous
  robot hands,'' \emph{Robotics and Autonomous Systems}, vol.~74, pp. 195--220,
  2015.

\bibitem{yuan2017gelsight}
W.~Yuan, S.~Dong, and E.~H. Adelson, ``Gelsight: High-resolution robot tactile
  sensors for estimating geometry and force,'' \emph{Sensors}, vol.~17, no.~12,
  p. 2762, 2017.

\bibitem{wang2021gelsight}
S.~Wang, Y.~She, B.~Romero, and E.~Adelson, ``Gelsight wedge: Measuring
  high-resolution 3d contact geometry with a compact robot finger,'' in
  \emph{2021 IEEE International Conference on Robotics and Automation
  (ICRA)}.\hskip 1em plus 0.5em minus 0.4em\relax IEEE, 2021, pp. 6468--6475.

\bibitem{taylor2022gelslim}
I.~H. Taylor, S.~Dong, and A.~Rodriguez, ``Gelslim 3.0: High-resolution
  measurement of shape, force and slip in a compact tactile-sensing finger,''
  in \emph{2022 International Conference on Robotics and Automation
  (ICRA)}.\hskip 1em plus 0.5em minus 0.4em\relax IEEE, 2022, pp.
  10\,781--10\,787.

\bibitem{romero2020soft}
B.~Romero, F.~Veiga, and E.~Adelson, ``Soft, round, high resolution tactile
  fingertip sensors for dexterous robotic manipulation,'' in \emph{2020 IEEE
  International Conference on Robotics and Automation (ICRA)}.\hskip 1em plus
  0.5em minus 0.4em\relax IEEE, 2020, pp. 4796--4802.

\bibitem{tippur2022design}
M.~H. Tippur, ``Design and manufacturing methods for a curved all-around
  camera-based tactile sensor,'' Ph.D. dissertation, Massachusetts Institute of
  Technology, 2022.

\bibitem{sun2022soft}
H.~Sun, K.~J. Kuchenbecker, and G.~Martius, ``A soft thumb-sized vision-based
  sensor with accurate all-round force perception,'' \emph{Nature Machine
  Intelligence}, vol.~4, no.~2, pp. 135--145, 2022.

\bibitem{padmanabha2020omnitact}
A.~Padmanabha, F.~Ebert, S.~Tian, R.~Calandra, C.~Finn, and S.~Levine,
  ``Omnitact: A multi-directional high-resolution touch sensor,'' in \emph{2020
  IEEE International Conference on Robotics and Automation (ICRA)}.\hskip 1em
  plus 0.5em minus 0.4em\relax IEEE, 2020, pp. 618--624.

\bibitem{she2020exoskeleton}
Y.~She, S.~Q. Liu, P.~Yu, and E.~Adelson, ``Exoskeleton-covered soft finger
  with vision-based proprioception and tactile sensing,'' in \emph{2020 IEEE
  International Conference on Robotics and Automation (ICRA)}.\hskip 1em plus
  0.5em minus 0.4em\relax IEEE, 2020, pp. 10\,075--10\,081.

\bibitem{liu2022gelsight}
S.~Q. Liu and E.~H. Adelson, ``Gelsight fin ray: Incorporating tactile sensing
  into a soft compliant robotic gripper,'' in \emph{2022 IEEE 5th International
  Conference on Soft Robotics (RoboSoft)}.\hskip 1em plus 0.5em minus
  0.4em\relax IEEE, 2022, pp. 925--931.

\bibitem{wilson2020design}
A.~Wilson, S.~Wang, B.~Romero, and E.~Adelson, ``Design of a fully actuated
  robotic hand with multiple gelsight tactile sensors,'' \emph{arXiv preprint
  arXiv:2002.02474}, 2020.

\end{thebibliography}

\end{document}